\pdfoutput=1

\documentclass[11pt]{article}

\usepackage[]{ACL2023}

\usepackage{times}
\usepackage{latexsym}

\usepackage[T1]{fontenc}

\usepackage[utf8]{inputenc}

\usepackage{microtype}

\usepackage{inconsolata}

\usepackage{diagbox}
\usepackage{graphicx}
\usepackage{verbatim}
\usepackage{multirow}
\usepackage{algorithmic}
\usepackage{longtable}
\usepackage{array}
\usepackage{subfigure}
\usepackage{stfloats}
\usepackage{balance}
\usepackage{multicol}
\usepackage{color}
\usepackage{epstopdf}
\usepackage{bm}
\usepackage{amsmath}
\usepackage{amssymb}
\usepackage{booktabs} 
\usepackage{epsfig}
\usepackage{enumitem}
\usepackage{cleveref}
\usepackage{arydshln}
\usepackage{balance}
\usepackage{xspace}
\usepackage{enumitem}
\usepackage{soul}

\newcommand{\keywordCode}[1]{{\small \texttt{#1}}}

\newcommand{\vpara}[1]{\vspace{0.05in}\noindent \textbf{#1 }}
\newcommand{\beq}[1]{\begin{equation}#1\end{equation}}
\newcommand{\beqn}[1]{
\begin{equation}
\begin{split}
#1
\end{split}
\end{equation}
}

\newcommand{\datasetname}{\mbox{\textsc{Today}}}
\newcommand{\matres}{\mbox{\textsc{Matres}}}
\newcommand{\tracie}{\mbox{\textsc{Tracie}}}
\newcommand{\stitle}[1]{\vspace{1ex}\noindent{\bf #1}}

\title{ \vspace*{-0.5in}

{{\small \hfill ACL'23}\\

\vspace*{.25in}} Generic Temporal Reasoning with Differential Analysis and Explanation
}

\author{Yu Feng, ~ Ben Zhou, ~ Haoyu Wang, ~ Helen Jin, ~ Dan Roth \\
{ University of Pennsylvania} \\
{\{fengyu1, xyzhou, why16gzl, helenjin, danroth\}@seas.upenn.edu}
}
\begin{document}
\maketitle

\begin{abstract}
Temporal reasoning is the task of predicting temporal relations of event pairs. While temporal reasoning models can perform reasonably well on in-domain benchmarks, we have little idea of these systems' generalizability due to existing datasets' limitations. In this work, we introduce a novel task named \datasetname{} that bridges this gap with \textbf{t}emp\textbf{o}ral \textbf{d}ifferenti\textbf{a}l anal\textbf{y}sis, which as the name suggests, evaluates whether systems can correctly understand the effect of incremental changes. Specifically, \datasetname{} introduces slight contextual changes for given event pairs, and systems are asked to 
tell how this subtle contextual change would affect relevant temporal relation distributions.
To facilitate learning, \datasetname{} also annotates human explanations. 
We show that existing models, including GPT-3.5, drop to random guessing on \datasetname{}, suggesting that they heavily rely on spurious information rather than proper reasoning for temporal predictions. 
On the other hand, we show that \datasetname{}'s supervision style and explanation annotations can be used in joint learning, encouraging models to use more appropriate signals during training and thus outperform across several benchmarks. \datasetname{} can also be used to train models to solicit incidental supervision from noisy sources such as GPT-3.5, thus moving us more toward the goal of generic temporal reasoning systems.

\end{abstract}
\section{Introduction}
\begin{figure}[ht]
\centering
\scalebox{0.48}{
	
	\includegraphics[width=0.98\textwidth]{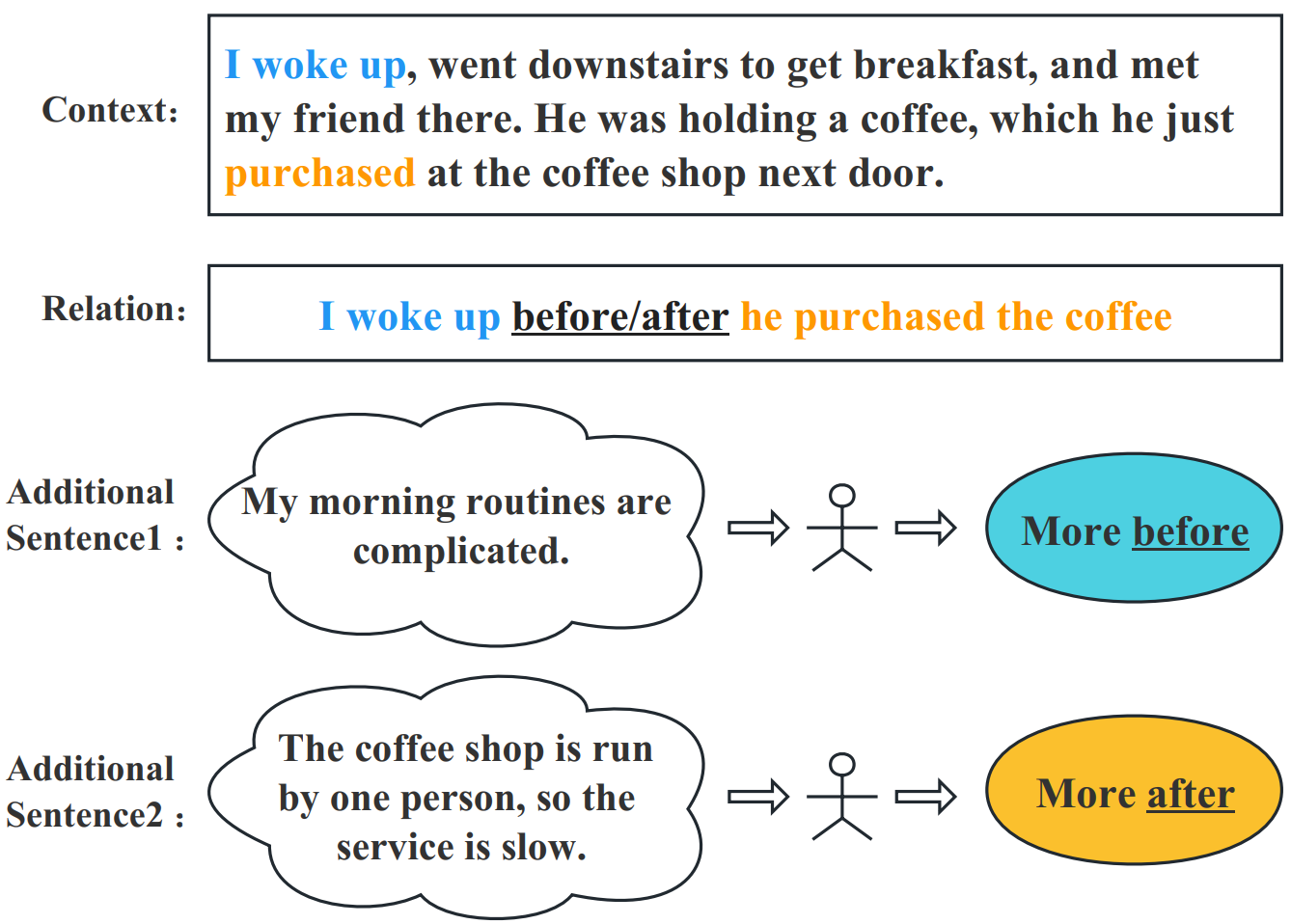}}
	\caption{\label{fig:main} A morning and coffee shop scenario example of temporal differential analysis. When adding the Additional Sentence 1 to the context, the temporal relation between the pair of events shifts towards \textbf{before}. Meanwhile, when adding the Additional Sentence 2,  the relation shifts towards \textbf{after}.}
\end{figure}
Temporal relation extraction \cite{pustejovsky2003timebank, chambers-etal-2014-dense} is traditionally viewed as an information extraction task, where a model uses explicit temporal signals such as ``happened before'' to identify the temporal order of events. While these models have contributed to many downstream pipelines, they are not enough for more complicated tasks such as timeline generation, where most event pairs do not come with explicit signals. These implicit temporal relation extractions \cite{zhou-etal-2021-temporal} thus require temporal reasoning, which relies on both common sense and semantic understanding of the context. In recent works, a popular approach to address these predictions is to finetune pre-trained language models (PLMs) with annotated supervision data. Unfortunately, existing temporal benchmarks \cite{pustejovsky2003timebank, cassidy-etal-2014-annotation, ning-etal-2018-multi} 
only annotate hard labels and ignore the fact that temporal labels can often be soft and nondeterministic. This approach allows models to exploit spurious signals and annotation artifacts easily for performance. For example, a model may learn to predict ``lunch'' before ``dinner'' regardless of the surrounding context, yet most existing benchmarks will not challenge such beliefs because most ``lunch'' annotations will happen to be before ``dinner.'' This is not always the case though, e.g. if the ``lunch'' and ``dinner'' were today's lunch and yesterday's dinner, and we know that yesterday's dinner must happen before today's lunch. This means that the current high performances of existing models may be misleading, and the community may actually possess an inaccurate perception of the models' capacity to generalize. 

In this work\footnote{Dataset and code are available at: \url{http://cogcomp.org/page/publication_view/1008}}, we bridge this evaluation gap with a novel benchmark that evaluates whether a temporal reasoning model is making the correct predictions for the right reasons by properly identifying potential alternatives (e.g., ``dinner'' can be before ``lunch'' under certain contexts). 
Our intuition is that a model with good temporal generalizability should be able to understand the effect of subtle context changes and \textit{explain} how the change will shift the temporal relation distribution of an event pair. To evaluate this, we propose the framework called \textbf{temporal differential analysis}.
Under this setting, we select event pairs where the temporal relation is not 100\% deterministic based on the context, meaning that both before/after relations are possible if additional information in regard to the context is given. Then, we annotate a hypothetical change in the form of an additional sentence added to the beginning of the context. As Fig.~\ref{fig:main} shows, this context change will shift the event pair's temporal relation distribution, making it either ``\textit{more before}'' or ``\textit{more after}''. Each hypothetical change is also annotated with human explanations of why the change affects the temporal relation. We collect 2,241 such instances with a rigorous human annotation pipeline and call the resulting dataset \datasetname{} (\textbf{t}emp\textbf{o}ral \textbf{d}ifferenti\textbf{a}l anal\textbf{y}sis).  

We find that models that achieve relatively high in-domain test performances are brittle and demonstrate minimal capabilities for differentiating subtle context changes that affect temporal relations. For example, the PatternTime model \cite{zhou-etal-2021-temporal} that achieves 77\% binary accuracy on \tracie{} \cite{zhou-etal-2021-temporal}
drops dramatically to 54\% on \datasetname{}, which is barely above random guessing.  
To mitigate this gap, we propose a general joint-learning technique that uses temporal explanations that \datasetname{} annotates. Specifically, we argue that explanations of temporal relations are an excellent proxy for understanding temporal reasoning. We show models trained with \datasetname{}'s task formulation and explanation annotation are better at perceiving cross-dataset supervision and achieve superior performances on multiple datasets with a single model.

We also find that while large language models (LLMs) are not good enough for temporal differential analysis, they do sometimes produce reasonable explanations for a given temporal relation. We design a pipeline that automatically collects supervision signals based on this finding. The pipeline starts with giving GPT-3.5 \cite{ouyang2022training} both an instance from \datasetname{} and a hypothetical temporal relation, and then uses GPT-3.5 to generate several explanations. 
Finally, we train an explanation verifier based on \datasetname{}'s human annotations, which selects the generated explanations that are more likely to be plausible. We show that adding such explanations from GPT-3.5 further boosts the performance across our benchmarks.

Our contributions are threefold: 1) We design a novel evaluation framework and collect a new dataset \datasetname{} that uses differential analysis to test whether systems can perform temporal reasoning with the right reasons; 2) We show that \datasetname{}'s supervision, especially the use of explanations, contributes toward a generic temporal reasoning model; 3) We use LLMs to generate pseudo explanations and filter these with a novel explanation verification system to show that such incidental supervision signals are helpful.
\section{Related Work}

\stitle{Temporal Reasoning Models.}
Significant effort has been devoted to temporal reasoning, a challenging task that requires models to recognize not only the connection between event mentions but also their contexts.
Several statistical learning models \cite{mani2007three, ning-etal-2017-structured,ning-etal-2018-cogcomptime} have been proposed to characterize events based on features and learn to predict the temporal relations.
Recently, data-driven temporal reasoning approaches \cite{trong2022selecting, wang2022extracting, liu2021discourse, mathur-etal-2021-timers, zhou-etal-2020-temporal, han-etal-2019-joint} have witnessed great improvement over these feature-based models on benchmarks and are generally built upon deep neural models to predict temporal labels in an end-to-end fashion.
Nevertheless, the lack of interpretability has made these neural models untrustworthy to be deployed in real-world applications \cite{yin-etal-2022-sensitivity}, especially in critical areas such as healthcare, finance, and government. 
The differential analysis approach to temporal reasoning first introduced in this paper provides a new paradigm for evaluating the interpretability and generalizability of temporal reasoning models.

\stitle{Temporal Relation Datasets.}
From different perspectives, multiple research projects have focused on constructing temporal reasoning benchmarks. 
A series of seminal datasets, TimeBank \cite{pustejovsky2003timebank}, TempEval 1-3 \cite{verhagen-etal-2007-semeval, verhagen-etal-2010-semeval, uzzaman-etal-2013-semeval}, 
\textsc{Matres} \cite{ning-etal-2018-multi} and so forth, have annotated on newswire articles for events and temporal relations between events.
\textsc{Torque} \cite{ning-etal-2020-torque} examines models' capability in temporal reasoning in reading comprehension. \textsc{Tracie} \cite{zhou-etal-2021-temporal} introduces a novel dataset that evaluates the degree to which systems understand implicit events.
However, none of these datasets annotate reasons to encourage generic temporal reasoning. 

\stitle{Explanations.} The community has been studying explanations and how they can help reasoning tasks such as question answering. Several models have been proposed \cite{rajani-etal-2019-explain, Latcinnik2020ExplainingQA, kumar-talukdar-2020-nile, ZRYR22}, as well as evaluation benchmarks that aim to test if existing systems can properly utilize explanations \cite{Camburu2018eSNLINL, aggarwal-etal-2021-explanations}. Our work is closely related to this line of effort as we attempt to build a proxy benchmark that can be automatically evaluated for temporal explanations. Recent findings on large language models have also inspired several works to use them as explanation generators \cite{wiegreffe-etal-2022-reframing, marasovic-beltagy-et-al-2022-feb}.

\section{Dataset}
\label{sec:dataset}

In this section, we introduce the evaluation framework and collection process of \datasetname{}.

\subsection{Task overview}
The \datasetname{} dataset and its overall framework are designed to evaluate systems' ability to make temporal predictions with plausible reasons. Existing datasets, including \matres, \textsc{Torque}, and \tracie, only annotate common event pairs that align with human common sense. In other words, if an event pair does not strongly imply a temporal relation (e.g. over 80\% confidence), it will not be annotated and tested on systems. This allows pre-trained language models with millions of parameters to exploit annotation artifacts and priors that do not necessarily hold in certain contexts. For example, we know ``lunch'' is usually before ``dinner'', but this also depends on if they are performed by the same subject, at the same location, and/or on the same day. Unfortunately, current models often memorize such relations as immutable facts, leading to prediction errors in instances that are less common in real life. This intuition inspires us to build a framework to evaluate how much spurious information and priors current models are using.

\vpara{Temporal Explanations.}
An ideal method to evaluate whether models are making predictions in the right way is to let them explain why a certain prediction is made and evaluate the faithfulness and plausibility of the explanations. However, such an evaluation framework is almost impossible to achieve with current progress in natural language processing, where the two main challenges are: 1) it is extremely difficult to collect gold explanations that are sufficient to cover any possible sets of explanations; and 2) it is impossible to evaluate system generations using existing summarization metrics automatically.

\vpara{Temporal Differential Analysis.}
Because of the aforementioned challenges in directly evaluating system explanations, we propose an alternative that is a close proxy to the ideal form, namely temporal differential analysis. The core of the temporal differential analysis is to check if models can correctly identify how a subtle change to the context may affect the temporal relations of a given event pair. The intuition behind this choice is two-fold: 1) it is much easier for both annotators and models to produce an explanation if they know which dimension to focus on; and 2) this provides a binary evaluation measure that is deterministic and trustworthy in terms of reflecting how much spurious information models are using. 

Specifically, our differential analysis process is defined below. Given an original context $\mathcal{C}$, event 1 $\mathcal{E}_1$ and event 2 $\mathcal{E}_2$,
we assume a gold distribution $\mathbb{D}=\{P_{before}, P_{after}, P_{same}\}$ on the temporal relation between $\mathcal{E}_1$ and $\mathcal{E}_2$ concerning $\mathcal{C}$, where $P_{before}, P_{after}, P_{same}$ are the probabilities of the temporal relation being before, after and simultaneous respectively, and the probabilities altogether sum to 1. We then annotate two additional sentences $\mathcal{AS}_{before}$ and $\mathcal{AS}_{after}$, where the temporal relation distribution between $\mathcal{E}_1$ and $\mathcal{E}_2$ with respect to $\mathcal{AS}_{before}+\mathcal{C}$ results in an increased $P_{before}$, while similarly the distribution using $\mathcal{AS}_{after}+\mathcal{C}$ as the context has a higher $P_{after}$.

Table~\ref{tb:example} shows an example instance of temporal differential analysis, where an additional sentence $\mathcal{AS}_{before}$ has an effect on the temporal relation between the two events and shifts the label distribution towards ``before''. We conducted a human pilot study for this formulation and found that it is easier to annotate and achieve substantial improvement over the explanation quality than to directly ask annotators to provide custom explanations for an event pair. We therefore adopt the former formulation and create our evaluation dataset \datasetname{} through a multi-stage annotation process as described below.

\begin{table}[t]
\newcolumntype{?}{!{\vrule width 1pt}}
\newcolumntype{C}{>{\centering\arraybackslash}p{40em}}
\centering 
\renewcommand\arraystretch{1.0}
\small{
\begin{tabular}{@{}l@{}}
\toprule
\textbf{Example} \\ \midrule
\textbf{Context $\mathcal{C}$}: \textcolor{blue}{Tim’s tooth was hurting like crazy. His dentist} \\ \textcolor{blue}{took a look around in his mouth. One of his teeth was rotten.} \\ \textcolor{blue}{Once the tooth was pulled, Tim felt fine.}\\ 
\midrule
\textbf{Additional Sentence 1 ($\mathcal{AS}_{before}$)}: \textcolor{teal}{Tim always met his } \\
\textcolor{teal}{dentist regularly.}\\
\midrule
\textbf{Event 1 ($\mathcal{E}_1$)}: \textcolor{orange}{Tim scheduled an appointment with his dentist.} \\
\textbf{Event 2 ($\mathcal{E}_2$)}: \textcolor{orange}{Tim's tooth started to hurt like crazy.} \\
\midrule
\textbf{Explanation ($Exp$)}: \textcolor{teal}{Some people maintain regular visits to} \\ \textcolor{teal}{a dentist. Tim is one of these individuals and may have} \\ \textcolor{teal}{ already scheduled a regular appointment with his dentist }\\
\textcolor{teal}{before his tooth started to hurt.}\\
\bottomrule
\end{tabular}
}
\caption{
	\label{tb:example} An example of temporal differential analysis, where $\mathcal{AS}$ shifts the temporal relation between $\mathcal{E}_1$ and $\mathcal{E}_2$ to be more ``before''. See \S \ref{sec:dataset} for more details.
}
\end{table}

\subsection{Dataset Construction}
Following the definition of the temporal differential analysis framework above, we collect a dataset to carry out the actual evaluation. Each instance in \datasetname{} contains a context $\mathcal{C}$, an event pair $\mathcal{E}_1$, $\mathcal{E}_2$, and an additional sentence of either $\mathcal{AS}_{before}$ or $\mathcal{AS}_{after}$. In addition, we also annotate a human explanation $Exp$ regarding why the additional sentence affects the temporal relation between the two events. \datasetname{} is constructed in three steps: 1) event pair generation, 2) additional sentence and explanation annotation, and 3) annotation verification and cleaning. We detail this pipeline below. 

\vpara{Generating $\mathcal{C}$ and $\mathcal{E}$.}
We randomly sample short stories from the ROCStories dataset~\cite{mostafazadeh-etal-2016-corpus} as the context $\mathcal{C}$. For each story, we use GPT-3.5 \footnote{We use GPT-3.5 text-davinci-002 for data generation throughout the work.} to generate an implicit event phrase based on an explicit event phrase selected by GPT-3.5 at the same time. An implicit event is an event that is not explicitly mentioned by the given context but is still inferable and relevant, e.g. Event 1 in Table~\ref{tb:example}. A sample prompt can be referred to in Appendix Table~\ref{tb:prompt2} to construct an event pair. We do this for two main reasons: 1) events that are not explicitly mentioned by the context provide more uncertainty so that the event pair does not come with a deterministic temporal relation decided by the context; 2) this is closer to the format of \tracie{}, which we aim to compare system performance changes with. 

\vpara{Crowdsourcing $\mathcal{AS}$ and $Exp$.}
After generating $\mathcal{C}$ and $\mathcal{E}$'s, we use Mechanical Turk to ask crowdsourcing annotators to write potential $\mathcal{AS}_{before}$ and $\mathcal{AS}_{after}$ with respect to the provided information. The guideline asks annotators to write additional sentences that can be added to the beginning of the context to prevent models from using text positional information. The annotator is also asked to explain why they wrote $\mathcal{AS}$ and why it affects the temporal relation distribution. We use this as $Exp$. We design an annotation interface that is intuitive and filled with examples, and at the same time, we require annotators to pass a rigorous qualification test to demonstrate a proper understanding. We list our interfaces and tests in Fig.~\ref{fig:mturk} and Table~\ref{tb:qual}.

\vpara{Annotation Verification.}
We employ an additional verification stage for the human-written instances from the previous step. We provide annotators with the formatted textual entailment instance and ask if the entailment label changes in the expected direction. We collect two individual verifications per instance, and the instances accepted by all annotators appear in the test set.

\subsection{Statistics}
We collect 1,000 instances agreed upon by all annotators as the evaluation set and construct a silver training set with the remaining 1,241 instances that do not have unanimous annotator agreements.
\section{Modeling}
\label{sec:model}
In this section, we show how to fully use \datasetname{}'s supervision signals (especially the explanations) to build a more generic temporal reasoning model. 

\vpara{Joint Learning.}\datasetname{} annotates temporal distribution shifts instead of absolute relations. This means that an instance may have a gold label ``before'' (i.e., the additional sentence $\mathcal{AS}$ makes the relation more ``before'' compared to the original context), yet the likelihood of ``after'' can still be higher, and the \textit{argmax} label will be ``after''. As a result, a model cannot sufficiently learn to predict absolute labels with only supervision signals from \datasetname{}. To mitigate this issue, we propose a joint learning model that requires joint supervision from a dataset that annotates hard labels for temporal relations, such as \matres{} or \tracie{}. 

\vpara{Modeling.} We adopt \tracie{}'s formulation~\cite{zhou-etal-2021-temporal} to format temporal reasoning 
into textual entailment and use a seq-to-seq
pre-trained language model as the base model.
Specifically, the input sequence consists of the premise, which is $\mathcal{AS} + \mathcal{C} + Exp$\footnote{$\mathcal{AS}$ and $Exp$ only apply for relative label instances, such as those in \datasetname{}.} in our case, as well as the hypothesis, which is $\mathcal{E}_1$ \keywordCode{starts [r]} $\mathcal{E}_2$. Here, $r$ is a hypothetical relation we plug into the hypothesis since systems are unaware of the gold label from the input sequence. The output sequence contains an entailment label, which is either \keywordCode{answer: positive} for entail or \keywordCode{answer: negative} for contradiction.

\vpara{Hard Label Instances.} As we note above, a system does not know the gold label when plugging in the hypothetical relation in the hypothesis. As a result, at learning time, we construct two entailment instances for a temporal relation instance with an absolute hard label. The first instance uses a hypothesis that is $\mathcal{E}_1$ \keywordCode{starts before} $\mathcal{E}_2$. We want the model to learn to output \keywordCode{answer: positive} for entail if the gold label is also ``before'', or \keywordCode{answer: negative} for contradiction if the gold label is ``after''. The second instance uses $\mathcal{E}_1$ \keywordCode{starts after} $\mathcal{E}_2$ as the hypothesis, where the output sequences are reversed compared to the first one. We use the regular cross-entropy loss for optimization and denote the loss as $\ell_{CE}$. At test time, we similarly construct two entailment instances for each event pair and
conduct a simple probability-based vote to infer 
a final ``before/after'' relation. 

\vpara{Relative Label Instances.} For instances that do not annotate absolute hard labels, we similarly construct two entailment instances for each event pair.
However, instead of 
using a cross-entropy loss to learn to output entailment labels, we employ a marginal ranking loss and ask the model to increase the probability of the entailment sequence if the plugged-in relation $r$ is the same as the gold label\footnote{Here ``gold label'' refers to the direction that $\mathcal{AS}$ shifts the temporal distribution to.} $r_g$, and vice versa. Specifically, we want: \footnote{For simplicity, we omit $Exp$ and $\mathcal{E}$ in the condition.}
\beq{
    \begin{cases}
        p(\mathrm{ent} |(\mathcal{AS}+\mathcal{C}), r)> p(\mathrm{ent} |\mathcal{C},r) & r = r_g \\ 
        p(\mathrm{con}|(\mathcal{AS}+\mathcal{C}),r)> p(\mathrm{con}|\mathcal{C},r) & r = \neg r_g
    \end{cases}
}
where $\mathrm{ent}$ and $\mathrm{con}$ represent entailment and contradiction respectively, 
and $\neg r_g$ is the opposite relation label of gold label $r_g$. 
The loss function we use can subsequently be written as:
\beqn{
\label{eq:marginrankingloss}
\ell_{MR} &= {\rm max}(0,\epsilon+p_{o_g}-p_{g}) \\&+ {\rm max}(0,\epsilon+p_{w}-p_{o_w})\\
p_{g} &= p(\mathrm{ent}|(\mathcal{AS}+\mathcal{C}),r_g) \\
p_{o_{g}} &= p(\mathrm{ent}|\mathcal{C},r_g)  \\
p_{w} &= p(\mathrm{ent}|(\mathcal{AS}+\mathcal{C}),\neg r_g)  \\
p_{o_{w}} &= p(\mathrm{ent}|\mathcal{C},\neg r_g)
}
\normalsize
where $\epsilon$ is a margin separating the logits. The actual probability of entailment is computed by the word logits in the output sequence of our model. 

\vpara{Aggregated Loss Function.} The final loss function we use for training considers both hard label instances and relative label instances, and is defined as follows:
\begin{equation}
\label{eq:loss}
\ell = \alpha \ell_{CE} + \ell_{MR}
\end{equation}
\normalsize
where $\alpha$ balances the two losses. As a result, we propose a general-purpose temporal reasoning model that can predict temporal relations for an event pair as well as probability changes for differential analysis as proposed in \datasetname{}.

\section{LLM Incidental Supervision}
\label{sec:incidental}

As we hypothesize and later show in \S\ref{sec:experiment}, human-annotated explanations greatly benefit generic temporal reasoning models, as they encourage models to learn to use the correct signals. However, it is extremely difficult and expensive to crowdsource such explanations for training purposes since collecting one instance costs \$1 on average. On the other hand, large language models (LLMs) can produce a large amount of generated explanations at a much cheaper cost. Unfortunately, these generated explanations are mostly unusable as they are simply model guesses based on 
textual correlations. 

In this section, we introduce a knowledge distillation method that combines the benefits of both human annotations and LLM generations by training verification models based on our seed annotation, which is then used to select generations more likely to be plausible. Compared to previous work \cite{wiegreffe-etal-2022-reframing}, we propose a verification system composed of multiple models that individually verify different aspects of automatically-generated explanations. We detail our pipeline below.

\subsection{Temporal Explanations from GPT-3.5}
We adopt the same event pair generation and context selection process as detailed in \S\ref{sec:dataset}. We design prompts as shown in Appendix Table~\ref{tb:prompt} and Table~\ref{tb:prompt1} that provide GPT-3.5 with contexts, event pairs, and temporal relations, and ask GPT-3.5 to generate additional sentences, how these sentences will change the temporal relations, and why. The prompt contains a few examples, which makes this setting few-shot. 

\subsection{Verification System}
\label{sec:gev}

\vpara{Similarity-based Filtering.}
We filter GPT-3.5 instances that use exact same sentences from the context as the additional sentence or repeat the event pairs and temporal relations as explanations. We use S-BERT~\cite{reimers-gurevych-2019-sentence} with a $0.95$ threshold to perform this filtering.

\vpara{General Explanation Verifier.}
We use the generic temporal relation model as proposed in \S\ref{sec:model} trained on \datasetname{} and an additional temporal relation dataset\footnote{Depending on the target task, this additional temporal relation dataset is different. We use \matres{} / \tracie{} / \matres{} + \tracie{} as the additional temporal relation dataset when evaluated on \matres{} / \tracie{} / All, respectively.} to verify if the generated additional sentence $\mathcal{AS}$ together with the explanation sentence $Exp$ shifts the temporal relation to the direction that it is supposed to.

\vpara{Additional Sentence Verifier.}
The general explanation verifier cannot sufficiently identify partial correctnesses of GPT-3.5 generations. For example, a generated instance may have a sub-optimal $\mathcal{AS}$ but convincing $Exp$, which could create deceptions. To address this, we train a separate $\mathcal{AS}$ verification model with \datasetname{} that does not use $Exp$ as input. We follow the same training scheme as \S\ref{sec:model}, and similarly, verify if the $\mathcal{AS}$ shifts the temporal relation as expected as our filtering criteria.

\vpara{Explanation Sentence Verifier.}
We also train a binary classification model to check the plausibility of $Exp$ individually. To generate negative $Exp$ instances, for each instance in the \datasetname{} training set with a given $\mathcal{AS}$, we ask GPT-3.5 to generate three possible explanation sentences. We use the one that is the least similar to the human-annotated $Exp$ according to S-BERT as the negative instance, which we denote as $Exp_{neg}$. We finetune the base seq-to-seq model with the positive and negative explanations and optimize the loss function as the negative log-likelihood of the positive explanation:
\beqn{
\ell^{E} &= -log\frac{e^{p_{pos}}}{e^{p_{pos}}+e^{p_{neg}}}\\
p_{pos} &= p(ent|(\mathcal{AS}+\mathcal{C},Exp_{human}),r_g) \\
p_{neg} &= p(ent|(\mathcal{AS}+\mathcal{C},Exp_{neg}),r_g)
}
We filter all GPT-3.5 generated instances whose explanation is deemed as negative by this binary classification model.

\section{Experiment}
\label{sec:experiment}
In this section, we conduct a series of experiments to show that 1) existing systems do not truly understand temporal relations, 2) \datasetname{} and incidental supervision signals partially address this issue, and 3) \datasetname{} motivates future work towards generic temporal reasoning. 

\subsection{Datasets, Metrics, and Settings}
We use our proposed dataset \datasetname{} as the main benchmark, as well as transferability results from two other temporal reasoning benchmarks \tracie{}~\cite{zhou-etal-2021-temporal} and \matres{}~\cite{ning-etal-2018-multi} to show that existing models fail to perform generic temporal reasoning while our proposal makes significant improvements. 
Following \citet{zhou-etal-2021-temporal}, all three datasets are processed as binary classification tasks by keeping instances that are originally annotated as either ``before'' or ``after''. As a result, we use binary accuracy as the metric. For \matres{}, we use only 1.5k (10\%) of the training instances to match the size of the other two datasets. Table~\ref{tab:datanum} summarizes data statistics.
We use $\epsilon=0.1$ in equation~\ref{eq:marginrankingloss} and $\alpha=10$ in equation~\ref{eq:loss}. All model training follows a standard textual entailment setup, uses default parameters, has the same number of steps, and averages from three random seeds. All training can be done with a single 48G-memory GPU within 5 hours.

\label{sec:datasetstats}
\begin{table}[ht]
\centering
\small
{
\scalebox{0.94}{
\begin{tabular}{lccccccc}
\toprule
Data &\#Train& \#Test & Relative-Label & Hard-Label\\
\cmidrule(lr){1-1}\cmidrule(lr){2-2}\cmidrule(lr){3-3}\cmidrule(lr){4-4}\cmidrule(lr){5-5}
\textsc{Today}&1,241&1,000&\checkmark&\\
\textsc{Tracie}&860&1,924&&\checkmark\\
\textsc{Matres}&1,500&1,322&&\checkmark\\
\bottomrule
\end{tabular}}
}
\caption{Statistics of the three datasets.} 
\label{tab:datanum}
\end{table}

\subsection{Baselines and Systems}
We report baseline performances of a state-of-the-art baseline PatternTime~\cite{zhou-etal-2021-temporal}, as well as GPT-3.5~\cite{brown2020language,ouyang2022training}. To show that \datasetname{} and other incidental supervision signals contribute to generic temporal reasoning, we use the T5-large model implemented by~\citet{wolf-etal-2020-transformers} as the base model and experiment with different supervision settings. We collect 5,000 GPT-3.5 generated instances in total, and 1,475 instances remain after our proposed verification models.

\begin{table*}[t]
\centering
\small
\begin{tabular}{lccccccc}
\toprule
Model (Train Data) & Loss & \tracie{} & \matres{} & \datasetname{} & \datasetname{} (gold exp.) & Average \\
\cmidrule(lr){1-1}\cmidrule(lr){2-2}\cmidrule(lr){3-3}\cmidrule(lr){4-4}\cmidrule(lr){5-5}\cmidrule(lr){6-6}\cmidrule(lr){7-7}\cmidrule(lr){8-8}
GPT-3.5 text-davinci-002 & FewShot&56.1&49.0&57.9&68.7&54.3 \\
GPT-3.5 text-davinci-003 & FewShot&52.3&50.1&59.0&70.0&53.8 \\
T5 (in-domain) & CE / MR & 66.2 & 81.2 & 52.9 & 55.7 & 66.8 \\
PatternTime & Distant & 77.0&73.0 &54.1&67.7&68.0\\
\cmidrule(lr){1-8}
T5 (O) & MR &50.6&49.8&52.9&55.7&51.1\\
T5 (O+G) & MR&55.4&52.3&55.0&66.5&54.2 \\
\cmidrule(lr){1-8}
T5 (M) & CE & 52.7 & 81.2 & 52.5& 57.5 & 62.1\\
T5 (M+O) & CE + MR & 51.5&81.7 &57.4&82.7&63.5\\
T5 (M+O+G) & CE + MR &49.9& 82.9&61.4&\textbf{82.9}& 64.8 \\
\cmidrule(lr){1-8}
T5 (T) & CE & 66.2 & 63.2 & 52.3&56.0 & 60.7\\
T5 (T+O) & CE + MR & 72.9 & 69.4 &59.9& 81.6 & 67.4\\
T5 (T+O+G) & CE + MR &73.5& 68.8& 62.1&82.0&68.1\\
\cmidrule(lr){1-8}
T5 (M+T) & CE & 66.2&82.0&52.5&58.5&66.9 \\
T5 (M+T+O) & CE + MR & 73.0 & 83.5 & 57.9& 77.8& 71.5\\
T5 (M+T+O+G) & CE + MR & 73.3&83.9&\textbf{63.2}&81.6 & 73.5\\
\cmidrule(lr){1-8}
PatternTime (M+T) & CE & 79.7 & 85.0 & 56.3 & 66.5 & 73.7 \\
PatternTime (M+T+O) & CE + MR & 79.8 & 85.8 & 60.9 & 82.2 & 75.5 \\
PatternTime (all) & CE + MR &\textbf{79.9}& \textbf{86.3}&62.9&82.3&\textbf{76.4}\\
\bottomrule
\end{tabular}
\caption{System performances under different supervision data and loss function settings across three binary temporal benchmarks. For simplicity, we use T to denote \tracie{} training data, and similarly M for \matres{}, O for \datasetname{} (ours), and G for GPT-3.5-generated incidental supervision. \datasetname{} (gold exp.) uses gold explanations during evaluation. \textit{Average} is averaged from \tracie{}, \matres{} and \datasetname{} accuracies. \textit{all} is equivalent to \textit{M+T+O+G}.}
\label{tab:maintable}
\end{table*}

\subsection{Main Results}
Table~\ref{tab:maintable} shows system performances under different supervision data and loss function settings across three binary temporal benchmarks, without generated explanations.

\vpara{Existing Work is Insufficient.}
We observe that GPT-3.5 is doing random guessing on all three benchmarks, suggesting that language model objectives alone are insufficient for temporal reasoning. On the other hand, PatternTime achieves mid-70s accuracy on \tracie{} and \matres{} but drops to random guessing on \datasetname{}. This suggests that biased supervision signals may improve on biased datasets,\footnote{Here, ``biased'' refers to datasets that align with natural distributions, such as \textit{drink coffee} is always before \textit{dinner}.} but not generic temporal reasoning. To further prove this point, we observe that T5 (M+T) jointly trained on \tracie{} and \matres{} does not improve much over T5 trained only on corresponding in-domain supervision (+0.4\% averaged accuracy), suggesting that previous temporal annotation styles do not motivate joint-learning nor generic temporal reasoning.

\vpara{Our Work Generalizes Better.}
On the contrary, we see that by simply using \datasetname{}'s moderate-sized 1k training instances, T5 (in-domain+O) improves 6.7\% on \tracie{}, and 0.5\% on \matres{}. When we add the incidental supervision instances from GPT-3.5 (filtered by \datasetname{}-supervised models in \S\ref{sec:incidental}, denoted as T5(in-domain+O+G) in Table~\ref{tab:maintable}), there is a 7.3\% improvement on \tracie{}, and 1.7\% on \matres{}. This is, on average, 4.5\% better than using \matres{} or \tracie{} as the supervision source. Moreover, \datasetname{} and incidental instances bring better joint learning efficiency and possibility, as we see a 6.7\% average accuracy improvement from T5(M+T+O+G) compared to T5's in-domain bests. If we use PatternTime\footnote{PatternTime also uses T5-large as the base model, and it does not use any in-domain annotation.} as the base model, we achieve a 76.4\% average accuracy which is the new state-of-the-art result of binary temporal relation classification across multiple datasets, and almost 10\% better than using T5 and in-domain supervision alone.

\vpara{Scaling and Improving LLMs is Inadequate.} We test the latest GPT-4 model \cite{OpenAI2023GPT4TR} on \datasetname{}, which gets 64.0\% accuracy, and 78.0\% with gold explanations.\footnote{We use the gpt-4-0314 checkpoint and chat API.} Even though GPT-4 is shown to significantly improve on many natural-language benchmarks over GPT-3.5, its improvement on \datasetname{} is relatively moderate, and it is only comparable with (if not worse than) our proposed model with less than a billion parameters. This shows that the advancement in large language models alone is insufficient to solve \datasetname{}, and more rigorous and controllable reasoning models are desirable for future works.

\subsection{Experiments with Generated Explanation}
\label{sec:inference} 
In Table~\ref{tab:maintable}, we see that explanations play an important role in generic temporal reasoning as \textit{PatternTime(all)} improves almost 20\% on \datasetname{} with the gold explanations. We, therefore, augment test instances with generated explanations on all three datasets. To utilize the existing explanation verification models proposed in \S\ref{sec:incidental}, we generate an additional sentence together with an explanation sentence. Specifically, for each possible relation direction of the event pair, we generate an additional sentence $\mathcal{AS}$ and an explanation sentence $Exp$ and then use explanation verifier models to select the $\mathcal{AS}$ and $Exp$ with the highest positive probability out of the two candidates. We use the same models and prompts described in \S\ref{sec:incidental}, and we show a sample of generated explanations in Table~\ref{tb:tracie}.\footnote {We use the given $\mathcal{AS}$ for \datasetname{}. We achieve this with the same prompt but only ask GPT-3.5 to generate an explanation sentence.}

Table~\ref{tab:generate_exp} shows model performances when augmented with generated explanations. There are improvements on all three datasets compared to the numbers in Table~\ref{tab:maintable}, with an average improvement of 1.0\% using T5 and 0.5\% using PatternTime. However, the overall performance is still suboptimal and the performance on \datasetname{} is far from when using gold explanations, which motivates future works on generating better explanations.

\begin{table}[t]
\centering
\small
{
\begin{tabular}{lccccccc}
\toprule
Model (Data) & T & M & \datasetname{} & Avg & $\bigtriangleup$ \\
\cmidrule(lr){1-1}\cmidrule(lr){2-2}\cmidrule(lr){3-3}\cmidrule(lr){4-4}\cmidrule(lr){5-5} \cmidrule(lr){6-6}
T5 (all) & 76.1& 84.4 & 63.1 & 74.5 & 1.0\\
PatternTime (all) & \textbf{80.5}  & \textbf{86.8} & \textbf{63.4} & \textbf{76.9} & 0.5 \\
\bottomrule
\end{tabular}
}
\caption{Model performances when augmented with generated explanations described in \S\ref{sec:inference}. T refers to \tracie{}, M refers to \matres{}, and Avg refers to Average. $\bigtriangleup$ shows the differences compared with Table \ref{tab:maintable}.} 
\label{tab:generate_exp}
\end{table}

\begin{table}[t]
\newcolumntype{?}{!{\vrule width 1pt}}
\newcolumntype{C}{>{\centering\arraybackslash}p{40em}}
\centering 
\renewcommand\arraystretch{1.0}
\small{
\begin{tabular}{@{}l@{}}
\toprule
\textbf{Example} \\ \midrule
\textbf{Context}: \textcolor{blue}{Jill studied all week for her math test. She stayed} \\ \textcolor{blue}{up studying the cold night before too. The morning of the} \\ \textcolor{blue}{ test, she woke up sick. But she went to school anyway. Jill's}\\ \textcolor{blue}{teacher allowed her to take the test at home.}  \\
\midrule
\textbf{Relation}: \textcolor{orange}{Jill's teacher trusted Jill \textbf{starts before} Jill's teacher} \\ \textcolor{orange}{allowed her to take the test at home.} \\
\midrule
\textbf{$\mathcal{AS}$}: \textcolor{teal}{Jill's teacher had always been impressed by her } \\
\textcolor{teal}{dedication to her studies.}\\
\midrule
\textbf{$Exp$}: \textcolor{teal}{The additional sentence implies jill's teacher allowed} \\ \textcolor{teal}{her to take the test at home because she trusted her and was}\\
\textcolor{teal}{impressed by her dedication.}\\
\bottomrule
\end{tabular}
}
\caption{
	\label{tb:tracie}An example of \tracie{} with generated explanations in \S\ref{sec:inference}. $\mathcal{AS}$ and $Exp$ are generated by GPT-3.5 and selected by our verification models described in \S\ref{sec:incidental}.
}
\end{table}

\begin{table}[t]
\centering
\small
{
\begin{tabular}{lccccccc}
\toprule
Ablation &\#GPT& T & M & \datasetname{} & Avg \\
\cmidrule(lr){1-1}\cmidrule(lr){2-2}\cmidrule(lr){3-3}\cmidrule(lr){4-4}\cmidrule(lr){5-5} \cmidrule(lr){6-6}
Ours&1,475&73.3&83.9&63.2&73.5\\
No Exp&1,867&73.7&83.5&61.2&72.8\\
No Addition&2,529&70.2&81.4&59.5&70.4\\
No General&2,079&71.0&81.8&59.5&70.8\\
More \#GPT&2,483&74.6&84.0&63.2&73.9\\
\bottomrule
\end{tabular}
}
\caption{Ablation study for LLM generated supervision. \textit{No Exp} does not use the explanation sentence verifier in \S\ref{sec:gev}, \textit{No Addition} does not use the additional sentence verifier, and \textit{No General} does not use the general verifier. \textit{More \#GPT} uses more verifier-filtered supervision instances (filtered
by three verifiers).} 
\label{tab:ablation}
\end{table}

\subsection{Ablation Studies and Human Analysis}
As shown in Table~\ref{tab:ablation}, we conduct ablation studies to better understand our incidental supervision signals. We see that the most rigorous setting with all three verifiers achieves the best performance with the fewest remaining instances. This suggests that all of our verifier models trained with \datasetname{} supervision are making positive contributions in selecting high-quality instances from GPT-3.5 generations.

We also see that using more incidental supervision instances verified by the verification models described in \S\ref{sec:incidental} can further enhance the model performance, suggesting a higher potential for using LLMs to generate supervision signals to empower smaller models. It also directs us to research the trade-off between model scaling and data scaling in temporal reasoning. 

We also conduct human analysis on the quality of the explanation sentences used in \datasetname{} and subsequent incidental supervision instances. We adopt the commonly used criteria for explanation~\cite{wiegreffe-marasovic-2021-review}, namely faithfulness (if an explanation implies the predicted label)~\cite{wiegreffe-pinter-2019-attention}, and plausibility (how well an explanation supports a predicted label)~\cite{deyoung-etal-2020-eraser}. We use Mechanical Turk to conduct human evaluation of the properties mentioned above. Given a differential analysis sample with an additional sentence and an explanation sentence towards a target temporal relation direction, we analyze faithfulness for the additional sentence by asking if it makes the temporal relation “more” toward the target relation and plausibility for the explanation sentence by asking if it explains why adding the differential content shifts the distribution toward the target relation. 

We show the experiment interfaces in Appendix Fig.~\ref{fig:eval} and present the results in Table~\ref{tab:human}. 
We randomly select 100 samples for each dataset for our human evaluation. For either faithfulness or plausibility, we collect two human evaluations for each sample. Only the sample that is valued as correct by both human annotators will be counted as a positive sample and we denote the total number of positive samples as the final score. We restrict each annotator to take 10 samples at most and there are 92 distinct annotators.
We see that \datasetname{}'s test set contains high-quality explanation annotations, which is expected from our rigorous agreement requirements. Our verification system improves both metrics for GPT-3.5 generated incidental supervision, which further demonstrates the effectiveness of the proposed verification models.

\begin{table}[t]
\centering
\small
{
\begin{tabular}{lccccccc}
\toprule
Data & Faithfulness& Plausibility \\
\cmidrule(lr){1-1}\cmidrule(lr){2-2}\cmidrule(lr){3-3}
\datasetname{} test&91&88\\
\datasetname{} train&79&68\\
GPT-3.5 distilled&80&67\\
GPT-3.5 random&57&55\\
\bottomrule
\end{tabular}
}
\caption{Human evaluation for faithfulness and plausibility of temporal differential analysis.
Faithfulness and Plausibility denote binary human evaluation results of the corresponding task. GPT-3.5 distilled refers to verifier-filtered GPT-3.5 data (filtered by three verifiers), and GPT-3.5 random refers to randomly sampled raw GPT-3.5 generated data. } 
\label{tab:human}
\end{table}

\section{Conclusion}
We introduce a novel differential analysis framework and dataset called \datasetname{} that interprets and evaluates if a temporal model can make correct predictions without using spurious information and biases. We show that existing temporal models' performances drop to random guessing on \datasetname{} due to model limitations and supervision biases. To address this issue, we propose to jointly train with \datasetname{} and its explanation annotations, resulting in improved performances on multiple temporal reasoning benchmarks, namely \tracie{} (+7\%), \matres{} (+3\%), and \datasetname{} (+10\%). We also demonstrate that \datasetname{} can be used to distill GPT-3.5 and automatically generate and filter incidental supervision instances with high-quality explanations, which further improves performances. Despite these advances, the gap in performance on \datasetname{} still motivates future work toward generic temporal reasoning. 
\section*{Limitations}
This work initially builds on human annotations, which are relatively expensive compared to simple model generations.
Due to such cost-related reasons, we do not include neutral contextual changes which are hard to annotate, and do not investigate the potential harms of annotated/generated language, e.g. harmful social biases. Throughout this work, we only use ROCStories as the source data, more diverse sources are reasonable for future work. We use T5 and GPT-3 architectures; however, there are more powerful architectures that could potentially improve our results.

Lastly, this work only focuses on generalizing temporal reasoning, which is a challenging yet relatively narrow task for large language models. Through pilot experiments, we find that similar task formulation, annotation schemes, and model structures can be applied to other tasks, such as natural language inference (NLI) and question answering (QA). A sample from the SNLI training set~\cite{bowman-etal-2015-large} using our formulation for explanation is shown in Table~\ref{tb:snli} in the Appendix.

\section*{Acknowledgements}
We thank the anonymous reviewers for their valuable feedback on this paper, as well as many others who provided constructive comments on the preprint. This work was supported by Contract FA8750-19-2-1004 with the US Defense Advanced Research Projects Agency (DARPA). Approved for Public Release, Distribution Unlimited. The views expressed are those of the authors and do not reflect the official policy or position of the Department of Defense or the U.S. Government.


\bibliographystyle{acl_natbib}

\appendix
\section{Appendix}
\begin{figure*}[b]
\centering
\scalebox{0.8}{
	
	\includegraphics[width=1\textwidth]{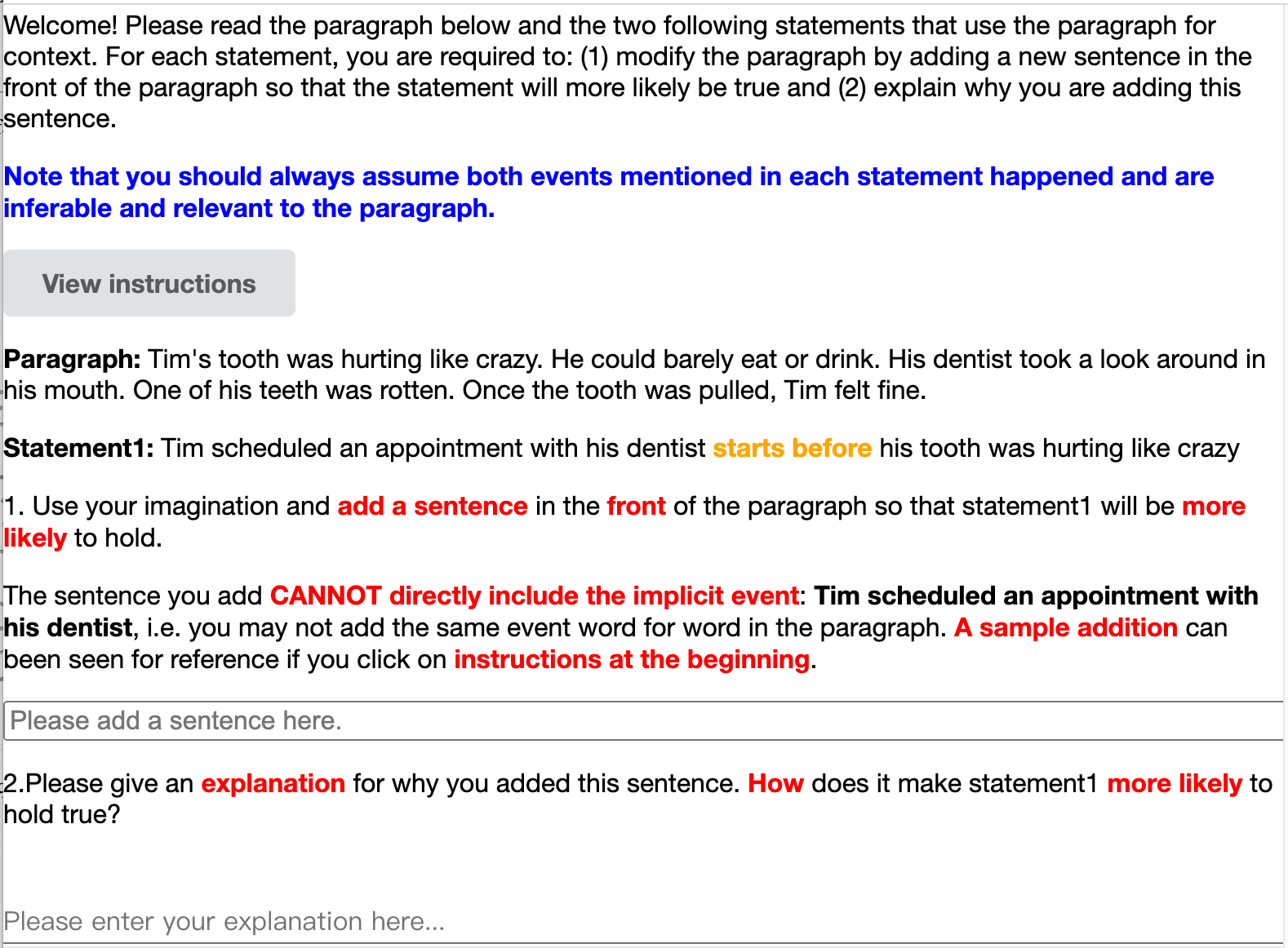}}
	\caption{\label{fig:mturk}The interface for differential analysis annotation. We only allow participants who have 90\% or more HITs acceptance rate, are located in the US, and pass our qualification task in Table \ref{tb:qual}. We also require annotators to spend at least 1.5 minutes for each instance (the hourly
salary is ~\$15). }
\end{figure*}

\begin{figure*}[ht]
\centering
\scalebox{0.8}{
	
	\includegraphics[width=1\textwidth]{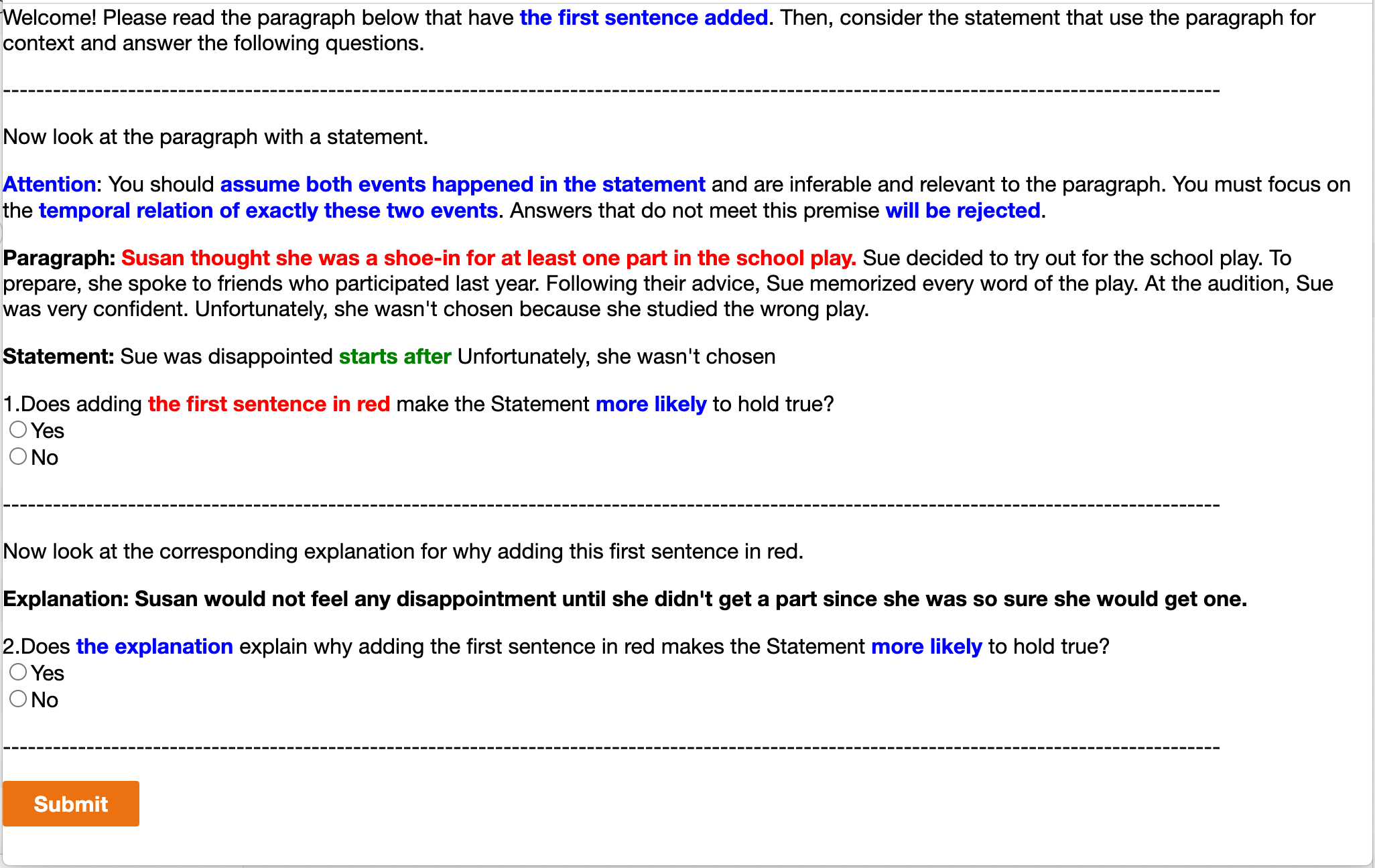}}
	\caption{\label{fig:eval}The interface for human evaluation. We only allow participants who have 98\% or more HITs acceptance rate, are located in the US, and pass our qualification task in Table \ref{tb:qual}. We also require annotators to spend at least 1 minute for each instance (the hourly
salary is ~\$15).}
\end{figure*}

\begin{table*}[ht]
\newcolumntype{?}{!{\vrule width 1pt}}
\newcolumntype{C}{>{\centering\arraybackslash}p{40em}}

\centering 
\renewcommand\arraystretch{1.0}
\small
\scalebox{0.95}{
\begin{tabular}{@{}l@{}}
\toprule
Let’s add a sentence to the first sentence of the context such that the hypothesis is more likely to hold true and explain why.\\ 
Context: \textcolor{blue}{
Tara always wanted jewelry. Her birthday was coming up. Test went to the store. He gave her a really nice necklace.}\\
 \textcolor{blue}{She adored him for the gift.
}\\
Hypothesis: \textcolor{orange}{Test was being a good friend \textbf{starts after} he give her a really nice necklace}\\

Add a sentence to the first sentence of the context such that the hypothesis is more likely to hold true and explain why.  \\
\textcolor{teal}{Test had a secret crush on a girl named Tara in the lower grade.} \\
\textcolor{teal}{Explanation: the fact that Test and Tara were in different grades implies that their relationship may not have been particularly close.}\\
\textcolor{teal}{However, Test's secret crush on Tara suggests that he paid close attention to her. By giving her the necklace, Test aimed to establish}\\
\textcolor{teal}{a stronger connection with Tara. }  \\
\#\#\# \\
Context: \textcolor{blue}{
Tara always wanted jewelry. Her birthday was coming up. Test went to the store. He gave her a really nice necklace.}\\
 \textcolor{blue}{She adored him for the gift.
}\\
Hypothesis: \textcolor{orange}{Test was being a good friend \textbf{starts before} he give her a really nice necklace}\\

Add a sentence to the first sentence of the context such that the hypothesis is more likely to hold true and explain why. \\
\textcolor{teal}{Test and Tara always hung out together.} \\
\textcolor{teal}{Explanation: normally people who hang out frequently are friends, and friends will send each other gifts on their birthdays.} \\
\#\#\# \\
Context: \textcolor{blue}{I have always been attracted to Hispanic men. That said, my first huge crush was on a Mexican. I was in love with}\\
\textcolor{blue}{him for two years. After two years, I realized I was wasting my time and idolizing him. Without any real sense of closure, I} \\
\textcolor{blue}{decided to pull my heart away.} \\
Hypothesis: \textcolor{orange}{I felt lonely \textbf{starts before} I decided to pull my heart away} \\
Add a sentence to the first sentence of the context such that the hypothesis is more likely to hold true and explain why.  
 \\\bottomrule
\end{tabular}
}
\caption{
	\label{tb:prompt} A sample prompt with an instance for two hypothetical changes to make the event pair's temporal relation "more before" or "more after".
}
\end{table*}

\begin{table*}[ht]
\newcolumntype{?}{!{\vrule width 1pt}}
\newcolumntype{C}{>{\centering\arraybackslash}p{40em}}

\centering 
\renewcommand\arraystretch{1.0}
\small
\scalebox{0.95}{
\begin{tabular}{@{}l@{}}
\toprule
Let’s add a sentence as the first sentence of the paragraph to let the statement more likely to hold true and explain why. \\ 
Paragraph: \textcolor{blue}{Tim's tooth was hurting like crazy. He could barely eat or drink. His dentist took a look around in his mouth. One of }\\
\textcolor{blue}{his teeth was rotten. Once the tooth was pulled, Tim felt fine.} \\
Statement: \textcolor{orange}{Tim scheduled an appointment with his dentist \textbf{starts after} his tooth started hurting like crazy} \\
Add what sentence as the first sentence of the paragraph and why is the statement more likely to hold true? \\
\textcolor{teal}{Tim's tooth was usually perfect, so he did not often go to see the dentist.} \\
\textcolor{teal}{This makes the statement true because it implies that Tim did not have regular appointments with his dentist and the reason why he} \\
\textcolor{teal}{scheduled an appointment with his dentist was that his tooth was hurting like crazy.}\\
\#\#\# \\
Paragraph: \textcolor{blue}{Tim's tooth was hurting like crazy. He could barely eat or drink. His dentist took a look around in his mouth. One of }\\
\textcolor{blue}{his teeth was rotten. Once the tooth was pulled, Tim felt fine.} \\
Statement: \textcolor{orange}{Tim scheduled an appointment with his dentist \textbf{starts before} his tooth started hurting like crazy} \\
Add what sentence as the first sentence of the paragraph and why is the statement more likely to hold true? \\
\textcolor{teal}{Tim always met his dentist regularly.} \\
\textcolor{teal}{This makes the statement true because it implies that Tim may have already scheduled regular appointments with his dentist before}  \\
\textcolor{teal}{his tooth started hurting like crazy.}  \\
\#\#\# \\
Paragraph: \textcolor{blue}{ Chuck was hanging out with some friends at a bar. They mentioned that they were moving soon. Chuck offered}\\
 \textcolor{blue}{to help them move their things. The team worked together and got the move done quickly. They were so grateful that they.
}\\
 \textcolor{blue}{invited him to stay for dinner.
 }\\
Statement: \textcolor{orange}{ Chuck wanted to be helpful \textbf{starts before} Chuck offered to help them move their things}\\
Add what sentence as the first sentence of the paragraph and why is the statement more likely to hold true? \\
\textcolor{teal}{Chuck is the kind of person that always wants to help out.} \\
\textcolor{teal}{This makes the statement true because it implies Chuck's wanted to help his friends move their things was because he is naturally}  \\
\textcolor{teal}{helpful.}  \\
\#\#\# \\
Paragraph: \textcolor{blue}{ Chuck was hanging out with some friends at a bar. They mentioned that they were moving soon. Chuck offered}\\
 \textcolor{blue}{to help them move their things. The team worked together and got the move done quickly. They were so grateful that they.
}\\
 \textcolor{blue}{invited him to stay for dinner.
 }\\
Statement: \textcolor{orange}{ Chuck wanted to be helpful \textbf{starts after} Chuck offered to help them move their things}\\
Add what sentence as the first sentence of the paragraph and why is the statement more likely to hold true? \\
\textcolor{teal}{Chuck often found himself reluctant to do thing, but grateful afterward that he did.} \\
\textcolor{teal}{This makes the statement true because if Chuck was reluctant, he might not have truly felt like being helpful until after he}  \\
\textcolor{teal}{offered to help and was grateful afterward.}  \\
\#\#\# \\
Paragraph: \textcolor{blue}{ I have always been attracted to Hispanic men. That said, my first huge crush was a Mexican. I was in love with}\\
 \textcolor{blue}{him for two years. After two years, I realized I was wasting my time and over-idolizing him. Without any real sense of closure, I
}\\
 \textcolor{blue}{decided to pull my heart away.
 }\\
Statement: \textcolor{orange}{I felt lonely \textbf{starts before} I decided to pull my heart away}\\
Add what sentence as the first sentence of the paragraph and why is the statement more likely to hold true?
 \\\bottomrule
\end{tabular}
}
\caption{
	\label{tb:prompt1} A sample prompt with two instances for two hypothetical changes to make the event pair's temporal relation "more before" or "more after".
}
\end{table*}

\begin{table*}[ht]
\newcolumntype{?}{!{\vrule width 1pt}}
\newcolumntype{C}{>{\centering\arraybackslash}p{40em}}

\centering 
\renewcommand\arraystretch{1.0}
\small
\scalebox{0.95}{
\begin{tabular}{@{}l@{}}
\toprule
Let's find out an event that is unmentioned but can be inferred from the context and the temporal relation between the two events\\
 are not deterministic. The new event should not be longer than ten words and include only one verb. \\ 
Context: \textcolor{blue}{
Tara always wanted jewelry. Her birthday was coming up. Test went to the store. He gave her a really nice necklace}\\
 \textcolor{blue}{She adored him for the gift.
}\\
What is an event that is unmentioned but has some role and can be inferred from the context? \\
\textcolor{teal}{Test was being a good friend} \\
\textcolor{teal}{It can be inferred from She adored him for the gift.} \\
\#\#\# \\
Context: \textcolor{blue}{Tim's tooth was hurting like crazy. He could barely eat or drink. His dentist took a look around in his mouth. One of }\\
\textcolor{blue}{his teeth was rotten. Once the tooth was pulled, Tim felt fine.}\\
What is an event that is unmentioned but has some role and can be inferred from the context? \\
\textcolor{teal}{Tim scheduled an appointment with his dentist} \\
\textcolor{teal}{It can be inferred from Tim's tooth was hurting like crazy.} \\
\#\#\# \\
Context: \textcolor{blue}{Lily went to a nice restaurant. She ordered a steak. To her dismay the steak was rare. Lily was rather upset. She had }\\
\textcolor{blue}{to send it back.}\\
What is an event that is unmentioned but has some role and can be inferred from the context?
 \\\bottomrule
\end{tabular}
}
\caption{
	\label{tb:prompt2} A sample prompt to generate an implicit event given the context.
}
\end{table*}

\begin{table*}[ht]
\newcolumntype{?}{!{\vrule width 1pt}}
\newcolumntype{C}{>{\centering\arraybackslash}p{40em}}

\centering 
\renewcommand\arraystretch{1.0}
\small
\scalebox{0.95}{
\begin{tabular}{@{}l@{}}
\toprule
Please read the paragraph below and the two following statements that use the paragraph for context.\\
 Use your imagination and add a sentence in the front of the paragraph so that the statement will be more likely to hold.  \\ 
The sentence you add CANNOT directly include the implicit event: Tim scheduled an appointment with his dentist. \\
 \midrule 
\textbf{Paragraph}: Tim's tooth was hurting like crazy. He could barely eat or drink. His dentist took a look around in his mouth. One of \\
his teeth was rotten. Once the tooth was pulled, Tim felt fine. \\
\textbf{Statement 1}: Tim scheduled an appointment with his dentist \textbf{starts after} his tooth was hurting like crazy.\\
\\
\textbf{Question 1.1}: Which modified paragraph do you think is the most suitable to make statement 1 more likely to hold?\\
$\circ$ \textbf{Tim ate a lot of spicy food.} Tim's tooth was hurting like crazy. He could barely eat or drink. His dentist took a look around in \\
his mouth. One of his teeth was rotten. Once the tooth was pulled, Tim felt fine. \\
$\circ$ \textbf{Tim didn't schedule an appointment with his dentist.} Tim's tooth was hurting like crazy. He could barely eat or drink. His\\
dentist took a look around in his mouth. One of his teeth was rotten. Once the tooth was pulled, Tim felt fine. \\
$\bullet$ \textbf{Tim's tooth was usually perfect, so he did not often go to see the dentist.} Tim's tooth was hurting like crazy. He could barely\\
eat or drink. His dentist took a look around in his mouth. One of his teeth was rotten. Once the tooth was pulled, Tim felt fine. \\
\midrule 
\textbf{Paragraph}: Tim's tooth was hurting like crazy. He could barely eat or drink. His dentist took a look around in his mouth. One of \\
his teeth was rotten. Once the tooth was pulled, Tim felt fine. \\
\textbf{Statement 2}: Tim scheduled an appointment with his dentist \textbf{starts before} his tooth was hurting like crazy. \\
\\
\textbf{Question 1.2}: Which modified paragraph do you think is the most suitable to make statement 2 more likely to hold? \\
$\circ$ \textbf{Tim scheduled an appointment with his dentist.} Tim's tooth was hurting like crazy. He could barely eat or drink. His dentist\\
took a look around in his mouth. One of his teeth was rotten. Once the tooth was pulled, Tim felt fine.\\
$\circ$ \textbf{Tim was looking for a dentist.} Tim's tooth was hurting like crazy. He could barely eat or drink. His dentist took a look around\\
in his mouth. One of his teeth was rotten. Once the tooth was pulled, Tim felt fine. \\
$\bullet$ \textbf{Tim always met his dentist regularly.} Tim's tooth was hurting like crazy. He could barely eat or drink. His dentist took a look\\
around in his mouth. One of his teeth was rotten. Once the tooth was pulled, Tim felt fine. \\
\midrule
\textbf{Question 2}: Do you understand that the additional sentence and the explanation you write down must make the statement more \\
likely to hold true and irrelevant explanation answers like "good" or merely copying any part of the paragraph will not be paid? \\
$\bullet$ Yes \\
$\circ$ No \\
\bottomrule
\end{tabular}
}
\caption{
	\label{tb:qual}Qualification test of differential analysis annotation. Participants can take the qualification test 3 times and only those who answer each question correctly can be allowed for annotation and evaluation tasks. 
}
\end{table*}

\begin{table*}[htb]
\newcolumntype{?}{!{\vrule width 1pt}}
\newcolumntype{C}{>{\centering\arraybackslash}p{40em}}

\centering 
\renewcommand\arraystretch{1.0}
\small
\scalebox{0.95}{
\begin{tabular}{@{}l@{}}
\toprule
\textbf{Gold answer} \\
\midrule
Let’s explain classification decisions.\\
\textcolor{blue}{A young boy wearing a tank-top is climbing a tree.}\\
question: \textcolor{orange}{A boy was showing off for a girl.}\\
true, false, or neither? \textcolor{teal}{neither}\\
why? \textcolor{teal}{A boy might climb a tree to show off for a girl, but he also might do it for fun or for other reasons.} \\
\midrule
\textbf{GPT-3.5 generated differential analysis explanation} \\
\midrule
Let’s add a sentence to the first sentence of the context such that the hypothesis is more likely to hold true and explain why. \\ 
Context: \textcolor{blue}{
A young boy wearing a tank-top is climbing a tree.}\\
Hypothesis: \textcolor{orange}{A boy \textbf{was} showing off for a girl.}\\
Add what sentence as the first sentence of the context and why is the hypothesis more likely to hold true? \\
\sethlcolor{pink}\hl{A boy was climbing a tree and he saw a girl watching him.} \\
\sethlcolor{pink}\hl{Because the paragraph is about a boy climbing a tree, the statement is more likely to be true if the boy is climbing the tree to}\\
\sethlcolor{pink}\hl{show off for the girl.} \\
\#\#\# \\
Context: \textcolor{blue}{
A young boy wearing a tank-top is climbing a tree.}\\
Hypothesis: \textcolor{orange}{A boy \textbf{was not} showing off for a girl.}\\
Add what sentence as the first sentence of the context and why is the hypothesis more likely to hold true? \\
\sethlcolor{pink}\hl{The boy was not showing off for a girl, he was merely climbing a tree.} \\
\sethlcolor{pink}\hl{The statement is more likely to hold true because if the boy was showing off for a girl, it would be implied that he was doing.}\\
\sethlcolor{pink}\hl{something to impress her. However, since the sentence states that he was merely climbing a tree, it is more likely that he was not.}\\
\sethlcolor{pink}\hl{doing it to show off for anyone.}
 \\\bottomrule
\end{tabular}
}
\caption{
	\label{tb:snli} A training sample from SNLI with its gold annotated explanations from \cite{wiegreffe-etal-2022-reframing} compared with the GPT-3.5 generated explanations (highlighted in pink) under our differential analysis formulation. 
}
\end{table*}

\end{document}